\documentclass[10pt,twocolumn,letterpaper]{article}
\pdfoutput=1
\usepackage{iccv}
\usepackage{times}
\usepackage{epsfig}
\usepackage{graphicx}
\usepackage{amsmath}
\usepackage{amssymb}
\usepackage{subcaption}
\usepackage{color}
\usepackage{booktabs}
\usepackage{multirow}

\usepackage[pagebackref=true,breaklinks=true,letterpaper=true,colorlinks,bookmarks=false]{hyperref}
\usepackage[linesnumbered,ruled]{algorithm2e} 
\iccvfinalcopy 

\def\iccvPaperID{2286} 

\ificcvfinal\fi
\begin{document}

\title{AutoFocus: Efficient Multi-Scale Inference}
\author{Mahyar Najibi  $^*$ 
\hspace{1.5cm}
 Bharat Singh \thanks{Equal Contribution}   \hspace{1.5cm} Larry S. Davis \\
 University of Maryland, College Park\\
{\tt\small \{najibi,bharat,lsd\}@cs.umd.edu}
}

\maketitle

\begin{abstract}
This paper describes AutoFocus, an efficient multi-scale inference algorithm for deep-learning based object detectors. Instead of processing an entire image pyramid, AutoFocus adopts a coarse to fine approach and only processes regions which are likely to contain small objects at finer scales. This is achieved by predicting category agnostic segmentation maps for small objects at coarser scales, called FocusPixels. FocusPixels can be predicted with high recall, and in many cases, they only cover a small fraction of the entire image. To make efficient use of FocusPixels, an algorithm is proposed which generates compact rectangular FocusChips which enclose FocusPixels. The detector is only applied inside FocusChips, which reduces computation while processing finer scales. Different types of error can arise when detections from FocusChips of multiple scales are combined, hence techniques to correct them are proposed. AutoFocus obtains an mAP of 47.9\% (68.3\% at 50\% overlap) on the COCO test-dev set while processing 6.4 images per second on a Titan X (Pascal) GPU. This is 2.5$\times$ faster than our multi-scale baseline detector and matches its mAP. The number of pixels processed in the pyramid can be reduced by 5$\times$ with a 1\% drop in mAP. AutoFocus obtains more than 10\% mAP gain compared to RetinaNet but runs at the same speed with the same ResNet-101 backbone.

\end{abstract}

\section{Introduction}
Human vision is foveal and active \cite{aloimonos1988active,findlay2003active}. The fovea, which observes the world at high-resolution, only corresponds to 5 degrees of the total visual field \cite{kolb1995simple}. Our lower resolution peripheral vision has a field of view of 110 degrees \cite{strasburger2011peripheral}. To find objects, our eyes perform saccadic movements which rely on peripheral vision \cite{kienzle2009center}. When moving between different fixation points, the region in between is simply ignored, a phenomenon known as saccadic masking \cite{breitmeyer1976implications,irwin1988visual,matin1974saccadic}. Hence, finding objects is an active process and the search time depends on the complexity of the scene. For example, locating a face in a portrait photograph would take much less time than finding every face in a crowded market. 

\begin{figure}
\centering
\includegraphics[width=0.82\linewidth]{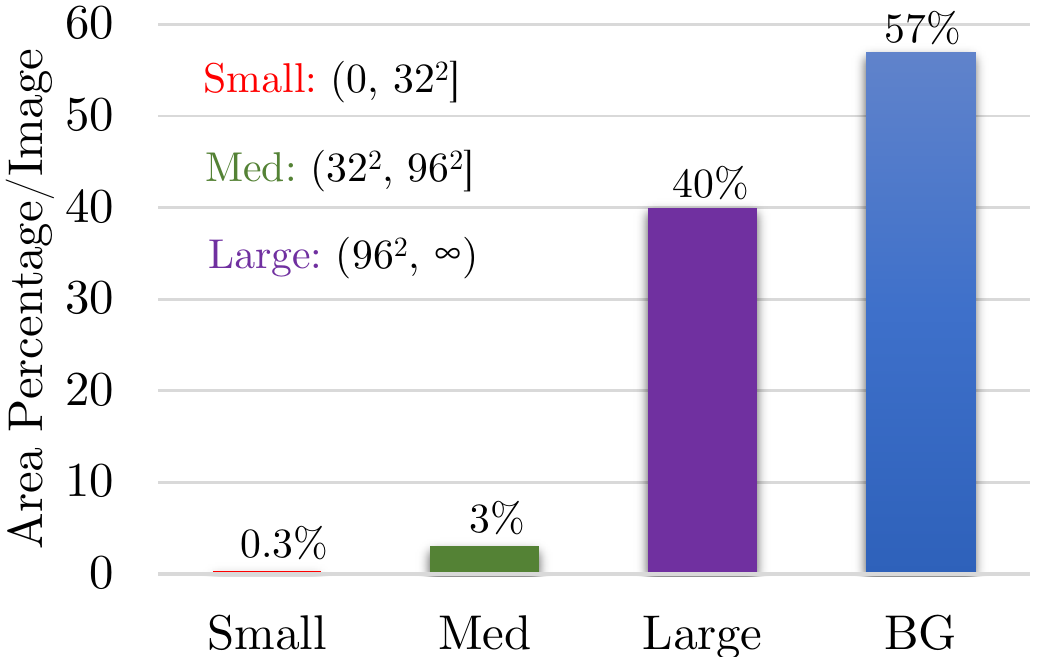}
\caption{Area of objects of different sizes and the background in the COCO validation set. Objects are divided based on their area (in pixels) into small, medium, and large.}
\label{fig:teaser}
\end{figure}

Adaptive processing, which is quite natural, brings several benefits. Many applications do not have real-time requirements and detectors are applied offline on billions of images/videos. Therefore, computational savings in a batch mode provide substantial monetary benefits. Examples include large-scale indexing of images and videos for visual search, APIs provided by cloud services, smart retail stores \etc. While there is work on image classification which performs conditional computation \cite{bengio2015conditional,liu2017dynamic,wu2017blockdrop}, modern object detection algorithms perform static inference and process every pixel of a multi-scale image pyramid to detect objects of different sizes \cite{singh2017analysis,sniper2018,peng2017megdet}. This is a very inefficient process as the algorithm spends equal energy at every pixel at different scales. 

To provide some perspective, we show the percentage of pixels occupied per image for different size objects in the COCO dataset in Fig \ref{fig:teaser}. Even though 40\% of the object instances are small, they only occupy 0.3\% of the area. If the image pyramid includes a scale of 3, then just to detect such a small fraction of the dataset, we end up performing 9 times more computation at finer-scales. If we add some padding around small objects to provide spatial context and only upsample these regions, their area would still be small compared to the resolution of the original image. So, when performing multi-scale inference, can we predict regions containing small objects from coarser scales?

If deep convolutional neural networks are an approximation of biological vision, it should be possible to localize object-like regions at lower resolution and recognize them by zooming on them at higher resolution - similar to the way our peripheral vision is coupled with foveal vision. To this end, we propose an object detection framework called {\em AutoFocus}, which adopts a coarse to fine approach and learns where to look in the next (larger) scale in the image pyramid. Thus, it saves computation while processing finer scales. This is achieved by predicting category agnostic binary segmentation maps for small objects, which we refer to as {\em FocusPixels}. A simple algorithm which operates on FocusPixels is designed to generate chips for the next image scale. AutoFocus only processes 20\% of the image at the largest scale in the pyramid on the COCO dataset, without any drop in performance. This can be improved to as little as 5\% with a 1\% drop in performance.

\section{Related Work}
Image pyramids \cite{witkin1984scale} and convolutional neural networks \cite{lecun1998gradient} are fundamental building blocks in the computer vision pipeline. Unfortunately, convolutional neural networks are not scale invariant. Therefore, for instance-level visual recognition problems, to recognize objects of different sizes, it is beneficial to rely on image pyramids \cite{singh2017analysis}. While efficient training solutions have been proposed for multi-scale training \cite{sniper2018}, inference on image pyramids remains a computational bottleneck which prohibits their use in practice. Recently, a few methods have been proposed to accelerate multi-scale inference, but they have only been evaluated under constrained settings like pedestrian/face detection or object detection in videos \cite{gao2018dynamic,song2018beyond,chen2018optimizing,shafiee2017fast,Liu_2018_CVPR,kang2017noscope}. In this work, we propose a simple and pragmatic framework to accelerate multi-scale inference for generic object detection which is evaluated on benchmark datasets.

Accelerating object detection has a long history in computer vision. The Viola-Jones detector \cite{viola2001rapid} is a classic example. It rejects easy regions with simple filters and spends more energy on promising object-like regions to accelerate the process. Several methods since then have been proposed to improve it \cite{bourdev2005robust,zhu2006fast,zhang2008multiple}. Prior to deep-learning based object detectors, it was common to employ a multi-scale approach for object detection \cite{viola2003detecting,dalal2005histograms,dollar2009integral,felzenszwalb2010object,dollar2010fastest,benenson2012pedestrian} and several effective solutions were proposed to accelerate detection on image pyramids. Common techniques involved approximation of features to reduce the number of scales \cite{dollar2010fastest,benenson2012pedestrian}, cascades \cite{bourdev2005robust,dollar2012crosstalk} or feature pyramids \cite{dollar2014fast}. Recently, feature-pyramids have been extensively studied and employed in deep learning based object detectors as the representation provides a boost in accuracy without compromising speed \cite{long2015fully,liu2016ssd,yang2016exploit,cai2016unified,lin2017feature,liu2017recurrent,najibi2017ssh,he2017mask,lin2018focal,liu2018path}. Although the use of image pyramids is common in challenge winning entries which primarily focus on performance \cite{he2016deep,dai2017deformable,peng2017megdet,liu2018path}, efficient detectors which operate on a single low-resolution image (\eg YOLO \cite{redmon2016you}, SSD \cite{liu2016ssd}, RetinaNet \cite{lin2018focal}) are commonly deployed in practice. This is because multi-scale inference on pyramids of high-resolution images is prohibitively expensive. 

AutoFocus alleviates this problem to a large extent and is designed to provide a smooth trade-off between speed and accuracy. It shows that it is possible to predict the presence of a small object at a coarser scale (referred to as FocusPixels) which enables avoiding computation in large regions of the image at finer scales. These are different from object proposals \cite{carreira2010constrained,uijlings2013selective,ren2015faster} where region candidates need to have a tight overlap with objects. Learning to predict FocusPixels is an easier task and does not require instance-level reasoning. AutoFocus shares the motivation with saliency and reinforcement learning based methods which perform a guided search while processing images \cite{itti1998model,hou2007saliency,liu2011learning,najibi2018towards,gonzalez2015active,mathe2016reinforcement,pirinen2018deep,jie2016tree,lu2016adaptive}, but it is  designed to predict small objects in coarser scales and they need not be salient.  

\begin{figure*}[!ht]
\centering
    \includegraphics[width=0.88\textwidth]{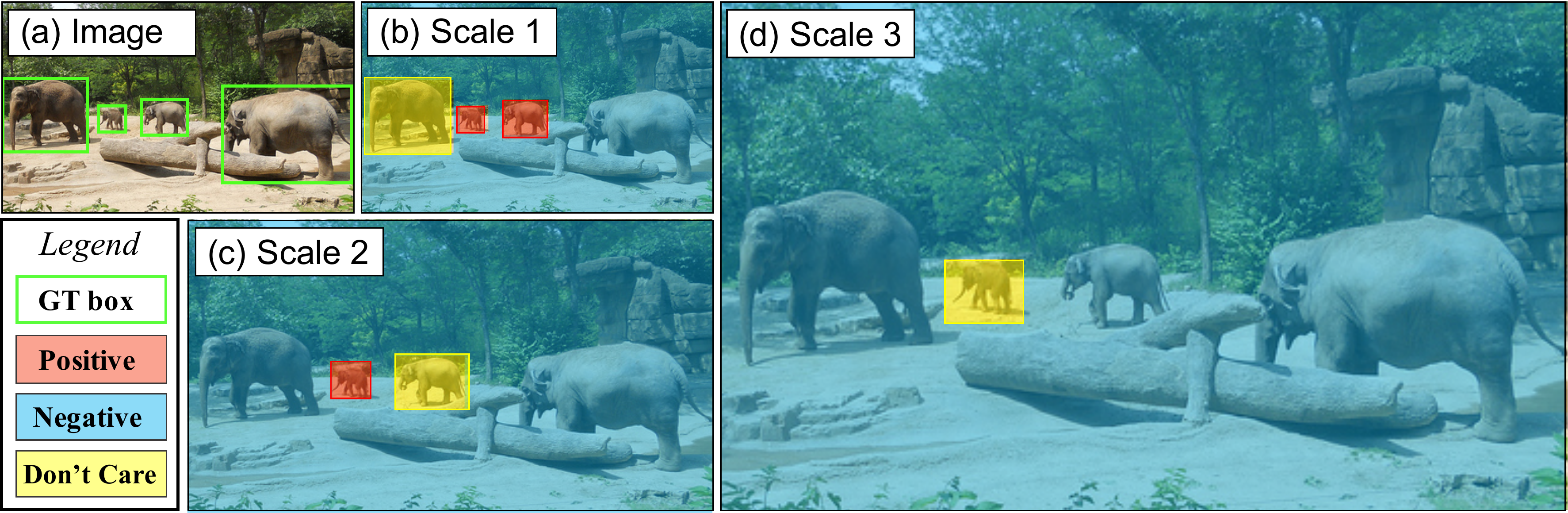}
\caption{The figure illustrates how FocusPixels are assigned at multiple scales of an image. At scale 1 (b), the smallest two elephants generate FocusPixels, the largest one is marked as background and the one on the left is ignored during training to avoid penalizing the network for borderline cases (see Sec. \ref{sec:focus_pixels} for assignment details). The labelling changes at scales 2 and 3 as the objects occupy more pixels. For example, only the smallest elephant would generate FocusPixels at scale 2 and the largest two elephants would generate negative labels.}
\label{fig:introPics}
\end{figure*}

\section{Background}

We provide a brief overview of SNIP, which is the multi-scale training and inference method described in \cite{singh2017analysis}. The core idea is to restrict the training samples to be in a pre-defined scale range which is appropriate for the input scale. For example, the detector is only trained on small objects at high resolution (larger scale) and large objects at low resolution (smaller scale). Because it is not trained on large objects at high resolution images, it is unlikely to detect them during inference as well. Rules are also defined to ignore large detections in high-resolution images during inference and vice-versa. Therefore, while merging detections from multiple scales, SNIP simply ignores large detections in high resolution images which contain most of the pixels.

Since the size of objects is known during training, it is possible to ignore large regions of the image pyramid by only processing appropriate context regions around objects. SNIPER \cite{sniper2018} showed that training on such low resolution {\em chips} with appropriate scaling does not lead to any drop in performance when compared to training on full-resolution images. If we can automatically predict these chips for small objects at a coarser scale, we may not need to process the entire high-resolution image during inference as well. But when these chips are generated during training, many object instances get cropped and their size changes. This did not hurt performance during training and can also be regarded as a data augmentation strategy. Unfortunately, if chips are generated during inference and an object is cropped into multiple parts, it would increase the error rate. So, apart from predicting where to look at the next scale, we also need to design an algorithm which correctly merges detections from chips at multiple scales.

\section{The AutoFocus Framework}
Classic features like SIFT \cite{lowe2004distinctive} / SURF \cite{bay2006surf}, combine two major components - the detector and the descriptor. The detector typically involved lightweight operators like Difference of Gaussians (DoG) \cite{marr1980theory}, Harris Affine \cite{mikolajczyk2004scale}, Laplacian of Gaussians (LoG) \cite{burt1987laplacian} \etc. The detector was applied on the entire image to find {\em interesting} regions. Therefore, the descriptor, which was computationally expensive, only needed to be computed for these interesting regions. This cascaded model of processing the image made the entire pipeline efficient. 

Likewise, the AutoFocus framework is designed to predict interesting regions in the image and discards regions which are unlikely to contain objects at the next scale. It zooms and crops only such interesting regions when applying the detector at successive scales. AutoFocus is comprised of three main components: the first learns to predict {\em FocusPixels}, the second generates {\em FocusChips} for efficient inference and the third merges detections from multiple scales, which we refer to as {\em focus stacking} for object detection. 

\subsection{FocusPixels}
\label{sec:focus_pixels}
FocusPixels are defined at the granularity of the convolutional feature map (like conv5). A pixel in the feature map is labelled as a FocusPixel if it has any overlap with a small object. An object is considered to be small if it falls in an area range (between 5 $\times$ 5 and 64 $\times$ 64 pixels in our implementation) in the resized chip (Sec. \ref{sec:chips}) which is input to the network . To train our network, we mark FocusPixels as positives. We also define some pixels in the feature map as invalid. Those pixels which overlap objects that have an area smaller or slightly larger than those defined as small are considered invalid (smaller than 5 $\times$ 5 or between 64 $\times$ 64 and 90 $\times$ 90). All other pixels are considered as negatives. AutoFocus is trained to generate high activations on regions which contain FocusPixels.

\begin{figure*}[!ht]
\centering
\includegraphics[width=\linewidth]{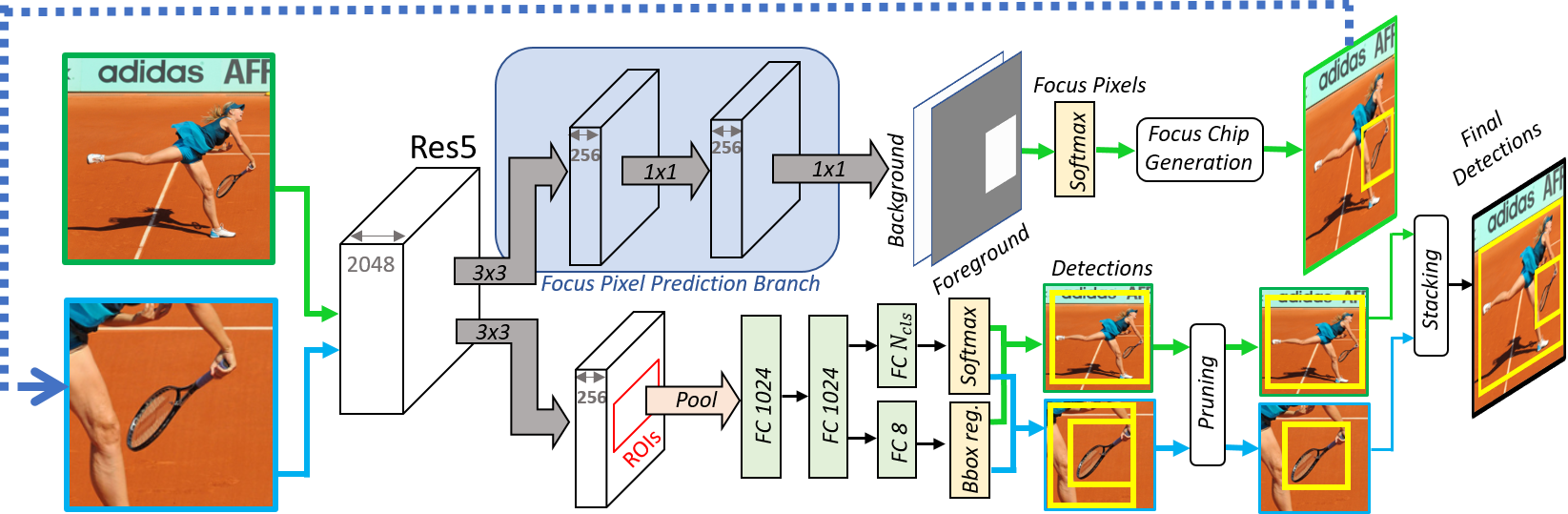}
\caption{The figure illustrates how AutoFocus detects a person and a racket in an image. The green borders and arrows are for inference at the original resolution. The blue borders and arrows are shown when inference is performed inside FocusChips. In the first iteration, the network detects the person and also generates a heat-map to mark regions containing small objects. This is depicted in the white/grey map - it is used to generate FocusChips. In the next iteration, the detector is then applied inside FocusChips only. Inside FocusChips, there could be detections for the cropped object present at the larger resolution. Such detections are pruned and finally valid detections are stacked across multiple scales.}
\label{fig:framework}
\end{figure*}

Formally, given an image of size $X \times Y$, and a fully convolutional neural network whose stride is $s$, then the labels $L$ will be of size $X' \times Y'$, where $X' = \lceil \frac{X}{s} \rceil$ and $Y' = \lceil \frac{Y}{s} \rceil$. Since the stride is $s$, each label $l \in L$ corresponds to $s \times s$ pixels in the image. The label $l$ is defined as follows,

\[
    l = 
\begin{cases}
    1,& IoU(GT, l) > 0, a < \sqrt{GTArea} < b \\
    -1,& IoU(GT, l) > 0, \sqrt{GTArea} < a  \\
    -1,& IoU(GT, l) > 0, b < \sqrt{GTArea} < c  \\
    0,              & \text{otherwise}
\end{cases}
\]
where $IoU$ is intersection over union of the $s\times s$ label block with the ground truth bounding box. $GTArea$ is the area of the ground truth bounding box after scaling. $a$ is typically 5, $b$ is 64 and $c$ is 90. If multiple ground-truth bounding boxes overlap with a pixel, FocusPixels ($l=1$) are given precedence. Since our network is trained on 512 $\times$ 512 pixel chips, the ratio between positive and negative pixels is around 10, so we do not perform any re-weighting for the loss. Note that during multi-scale training, the same ground-truth could generate a label of 1, 0 or -1 depending on how much it has been scaled. The reason we regard pixels for medium objects as invalid ($l=-1$) is that the transition from small to large objects is not visually obvious. Extremely small objects in each scale are also marked as invalid because after the early down-sampling operations, the network does not have sufficient information to make a correct prediction about them at that particular scale. The labelling scheme is visually depicted in Fig \ref{fig:introPics}. For training the network, we add two convolutional layers (3$\times$3 and 1$\times$1) with a ReLU non-linearity on top of the conv5 feature-map. Finally, we have a binary softmax classifier to predict FocusPixels, shown in Fig \ref{fig:framework}.

\subsection{FocusChip Generation}
\label{sec:chips}
During inference, we mark those pixels $\mathcal{P}$ in the output as FocusPixels, where the probability of foreground is greater than a threshold $t$, which is a parameter controlling the speed-up and can be set with respect to the desired speed accuracy trade-off. This generates a number of connected components $\mathcal{S}$. We dilate each component with a filter of size $d \times d$ to increase contextual information needed for recognition. After dilation, components which become connected are merged. Then, we generate chips $\mathcal{C}$ which enclose these connected components. Note that chips of two connected components could overlap. As a result, these chips are merged and overlapping chips are replaced with their enclosing bounding-boxes. Some connected components could be very small, and may lack the contextual information needed to perform recognition. Many small chips also increase fragmentation which results in a wide range of chip sizes. This makes batch-inference inefficient. To avoid these problems, we ensure that the height and width of a chip is greater than a minimum size $k$.  This process is described in Algorithm \ref{alg:chip_generation}. Finally, we perform multi-scale inference on an image pyramid but successively prune regions which are unlikely to contain objects. 

\begin{algorithm}[t!]
    \caption{FocusChip Generator}
        \label{alg:chip_generation}
	\SetKwInOut{Input}{Input}\SetKwInOut{Output}{Output}
    \Input{Predictions for feature map $\mathcal{P}$, threshold $t$, dilation constant $d$, minimum size of chip $k$} 
    \Output{Chips $\mathcal{C}$}
    \BlankLine
        Transform $\mathcal{P}$ into a binary map using the threshold $t$ \\
        Dilate $\mathcal{P}$ with a $d\times d$ filter\\
        Obtain a set of connected components $\mathcal{S}$ from $\mathcal{P}$\\
        Generate enclosing chips $\mathcal{C}$ of size $> k$ for each component in $\mathcal{S}$\\
        Merge chips $\mathcal{C}$ if they overlap\\
 	\Return{Chips $\mathcal{C}$}
 	
\end{algorithm}

\begin{figure*}[!ht]
\centering
\includegraphics[width=1\linewidth]{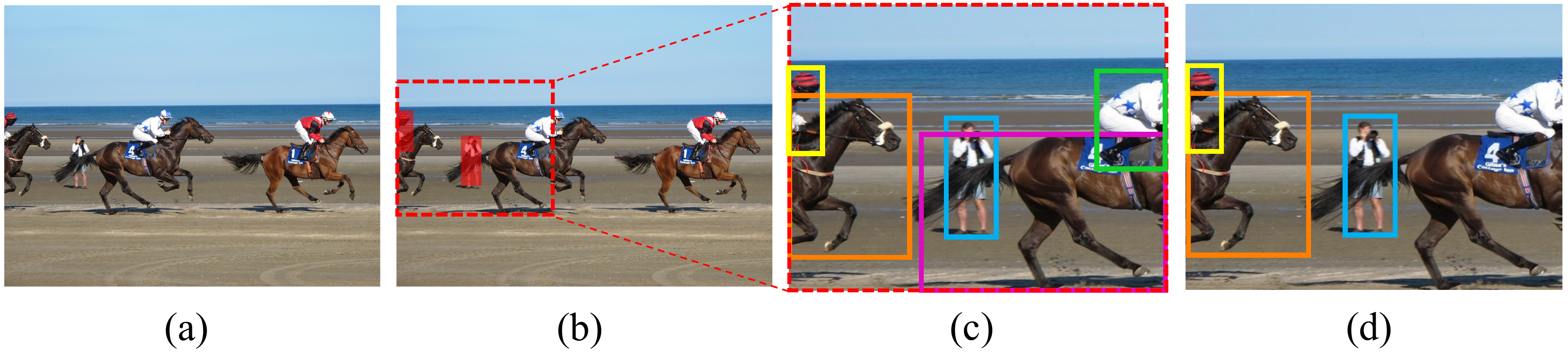}
\caption{Pruning detections while FocusStacking. (a) Original Image (b) The predicted FocusPixels and the generated FocusChip (c) Detection output by the network (d) Final detections for the FocusChip after pruning.}
\label{fig:FocusStacking}
\end{figure*}

\subsection{Focus Stacking for Object Detection}
One issue with such cascaded multi-scale inference is that some detections at the boundary of the chips can be generated for cropped objects which were originally large. At the next scale, due to cropping, they could become small and generate false positives, such as the detections for the horse and the horse rider on the right, shown in Fig \ref{fig:FocusStacking} \textcolor{red}{(c)}. To alleviate this effect, Step 2 in Algorithm \ref{alg:chip_generation} is very important. Note that when we dilate the map $\mathcal{P}$ and generate chips, this ensures that no {\em interesting} object at the next scale would be observed at the boundaries of the chip (unless the chip shares a border with the image boundary). Otherwise, it would be enclosed by the chip, as these are generated around the dilated maps. Therefore, if a detection in the zoomed-in chip is observed at the boundary, we discard it, even if it is within valid SNIP ranges, such as the horse rider eliminated in Fig \ref{fig:FocusStacking} \textcolor{red}{(d)}. 

There are some corner cases when the detection is at the boundary (or boundaries $x$, $y$) of the image. If the chip shares one boundary with the image, we still check if the other side of the detection is completely enclosed inside or not. If it is not, we discard it, else we keep it. In another case, if the chip shares both the sides with the image boundary and so does the detection, then we keep the detection.

Once valid detections from each scale are obtained using the above rules, we merge detections from all the scales by projecting them to the image co-ordinates after applying appropriate scaling and translation. Finally, Non-Maximum Suppression is applied to aggregate the detections. The network architecture and an example of multi-scale inference and focus stacking is shown in Fig \ref{fig:framework}.

\section{Datasets and Experiments}
We evaluate AutoFocus on the COCO \cite{lin2014microsoft} and the PASCAL VOC \cite{everingham2010pascal} datasets. As our baseline, we use the SNIPER detector\footnote{\url{http://www.github.com/mahyarnajibi/SNIPER}} \cite{sniper2018} which obtains an mAP of 47.9\% (68.3\% at 50\% overlap) on the COCO test-dev set and 47.5\% (67.9\% at 50\% overlap) on the COCO validation set. We add the fully convolutional layers for AutoFocus which predict the FocusPixels. No other changes are made to the architecture or the training schedule. We use Soft-NMS \cite{bodla2017soft} at test-time for Focus Stacking with $\sigma = 0.55$. Following SNIPER \cite{sniper2018}, the resolutions used for the 3 scales at inference are S1=$(480,512)$, S2=$(800,1280)$, and S3=$(1400,2000)$. The first resolution is the minimum size of a side and the second one is the maximum in pixels. The scales corresponding to these resolutions are referred to as scales 1, 2 and 3 respectively in the following sections. 

\begin{figure}
\centering
\includegraphics[width=0.6\linewidth]{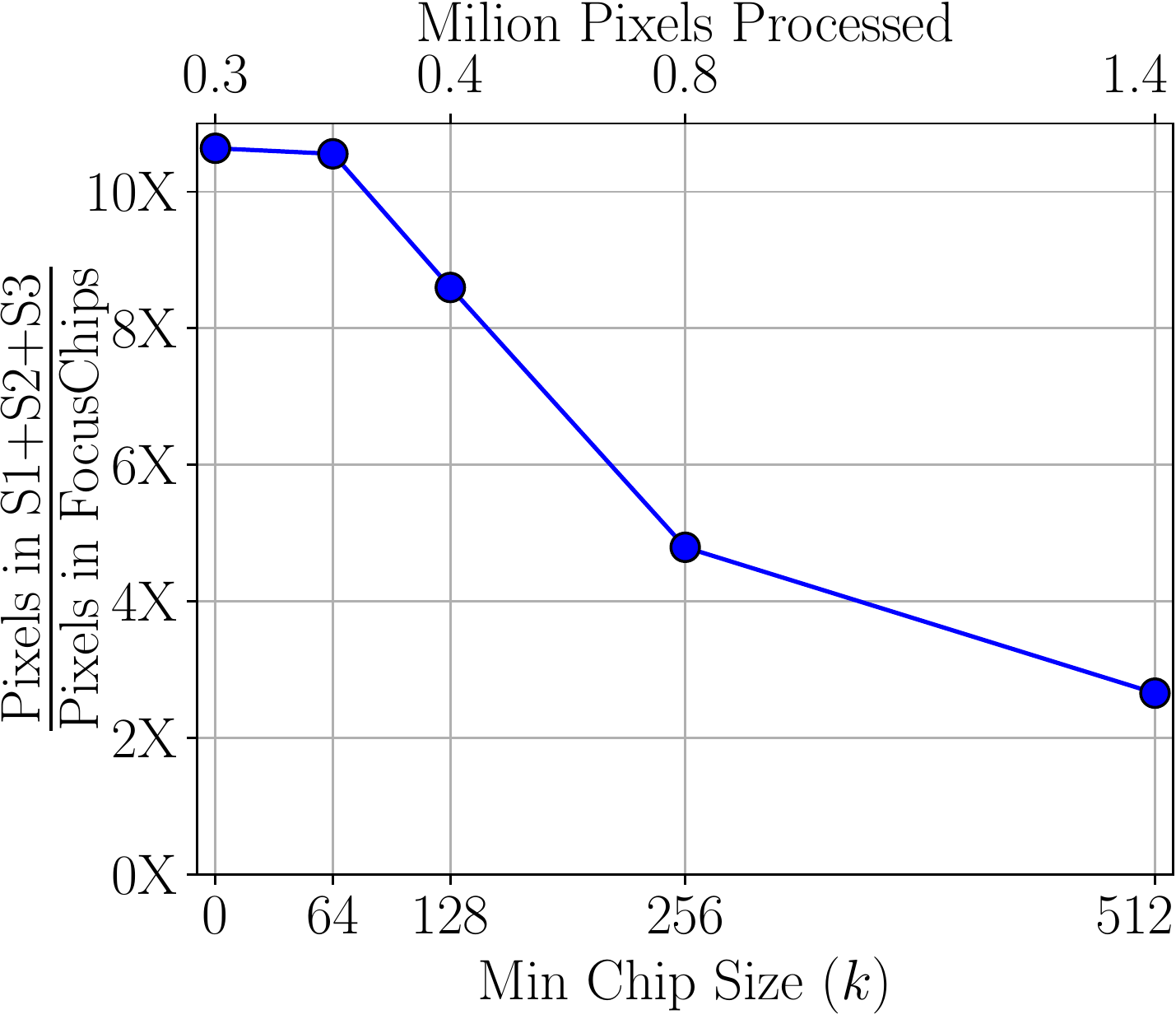}
\caption{Upper-bound on the speed-up using FocusChips generated from optimal FocusPixels.}
\label{fig:upperbound}
\end{figure}
Since FocusChips of different size are generated, we group chips which are of similar size and aspect ratio to achieve a high batch inference throughput. In some cases, we need to perform padding when performing batch inference, which can slightly change the number of pixels processed per image. For large datasets, this overhead is negligible as the number of groups (for size and aspect ratio) can be increased without reducing the batch size. 

\begin{figure*}
    \centering
    \begin{subfigure}[t]{0.261\linewidth}
    \includegraphics[width=\linewidth]{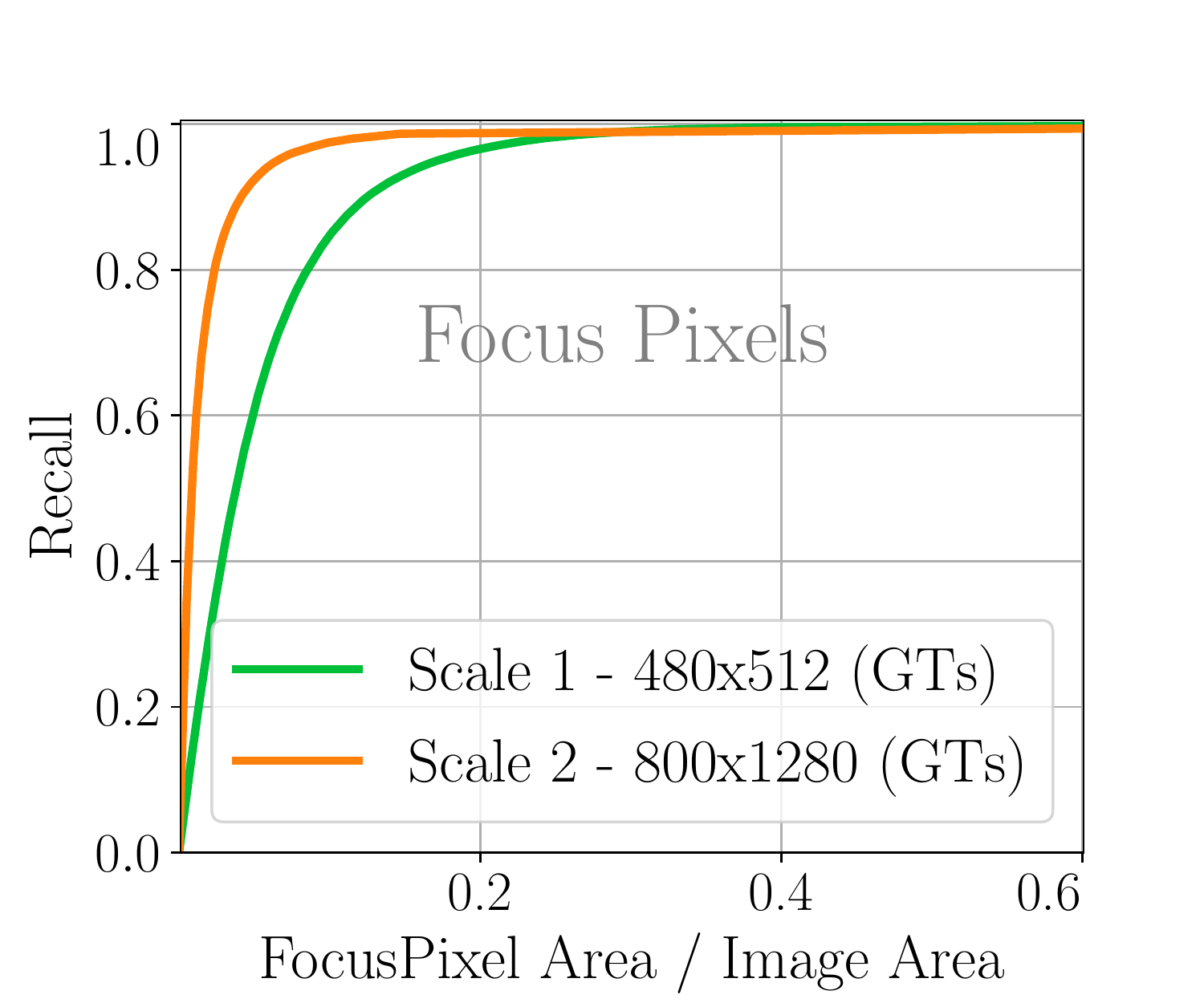}
    \caption{}
    \end{subfigure}
    \begin{subfigure}[t]{0.241\linewidth}
    \includegraphics[width=\linewidth]{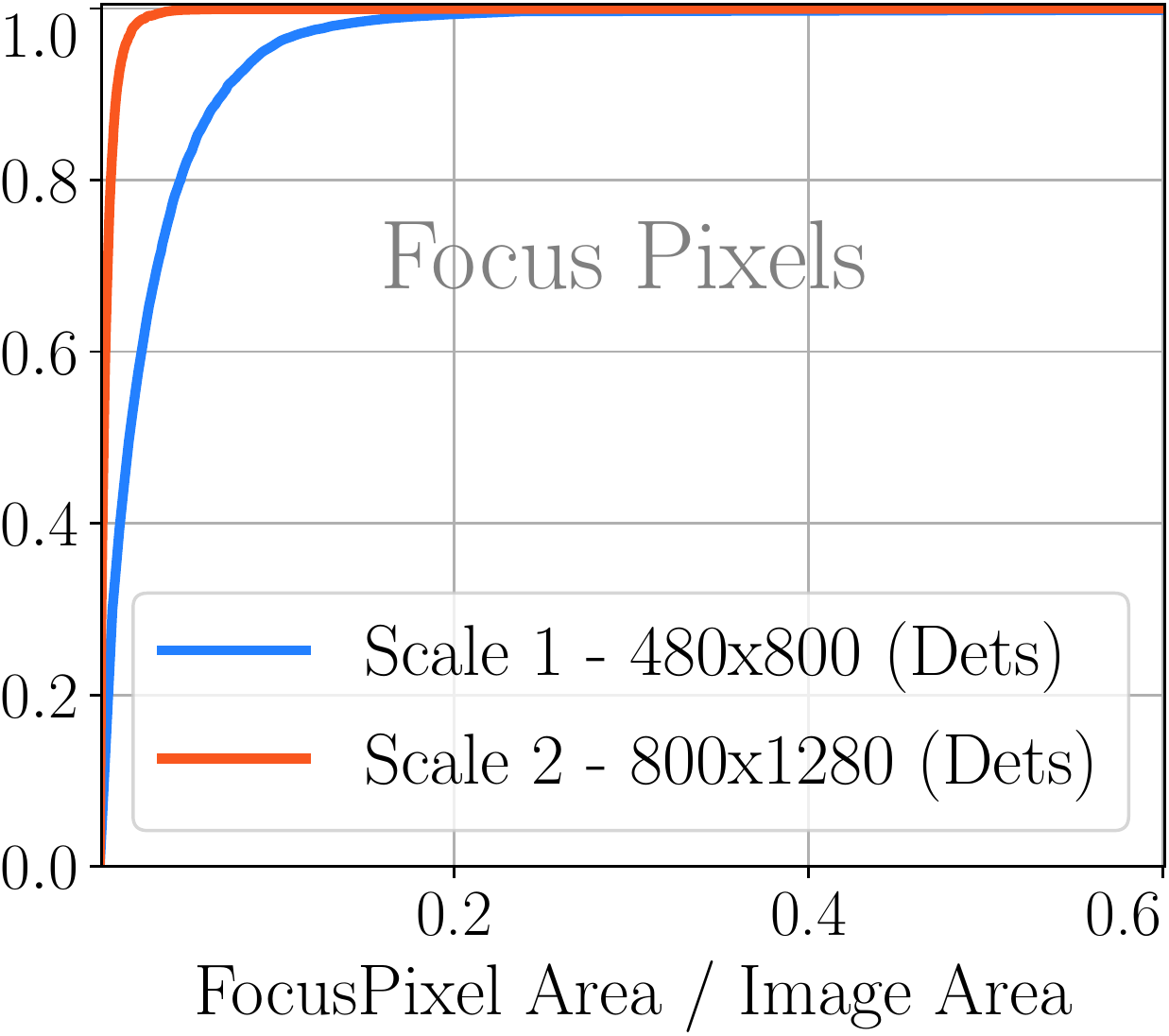}
    \caption{}
    \end{subfigure}
      \begin{subfigure}[t]{0.241\linewidth}
    \includegraphics[width=\linewidth]{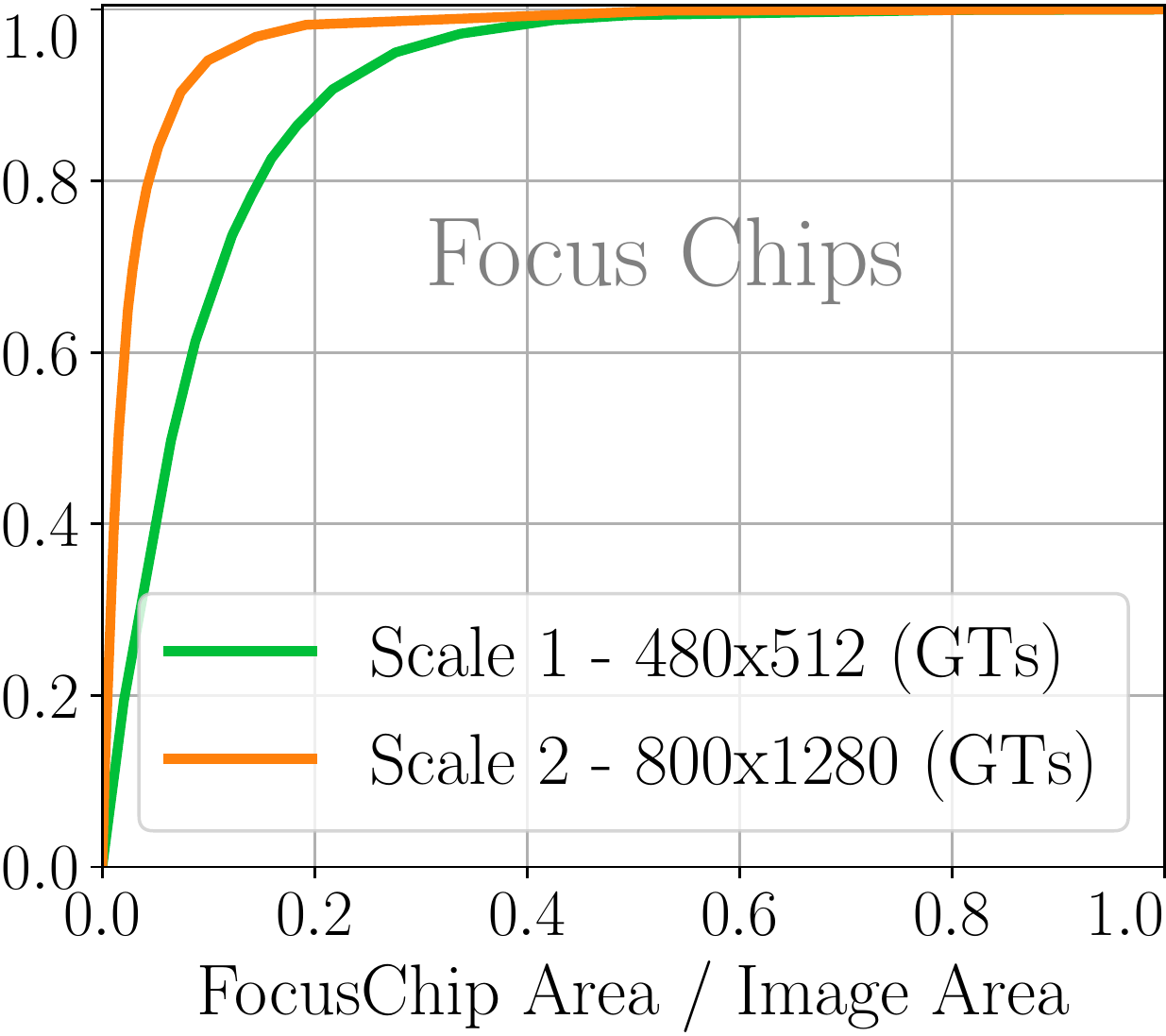}
    \caption{}
    \end{subfigure}
    \begin{subfigure}[t]{0.241\linewidth}
    \includegraphics[width=\linewidth]{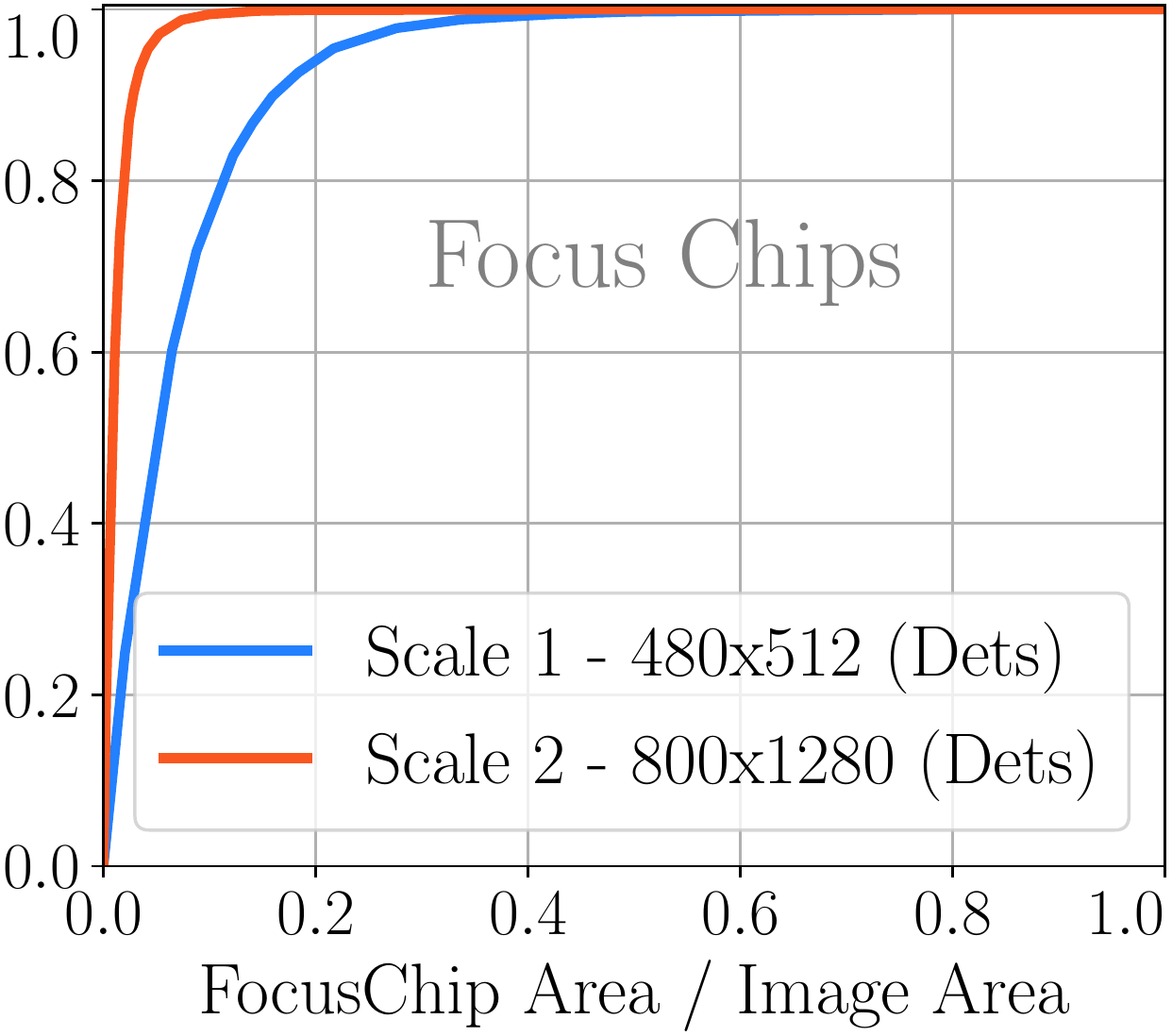}
    \caption{}
    \end{subfigure}
    \caption{Quality of the FocusPixels and FocusChips. The x-axis represents the ratio of the area of FocusPixels or FocusChips to that of the image. The y-axis changes as follows, (a) FocusPixel recall is computed based on the GT boxes (b) FocusPixel recall is computed using the confident detections (c) FocusChip recall is computed based on the GT boxes (d) FocusChip recall is computed based on the confident detections.}
    \label{fig:my_label}
\end{figure*}

\subsection{Stats for FocusPixels and FocusChips}
 
In high resolution images (scale 3), the percentage of FocusPixels is very low (\ie $\sim4$\%). So, ideally a very small part of the image needs to be processed at high resolution. Since the image is upsampled, the FocusPixels projected on the image occupy an area of $63^2$ pixels on average (the highest resolution images have an area of $1602^2$ pixels on average). At lower scales (like scale 2), although the percentage of FocusPixels increases to $\sim11\%$, their projections only occupy an area of $102^2$ pixels on average (each image at this scale has an average area of $940^2$ pixels). After dilating FocusPixels with a kernel of size 3 $\times$ 3, their percentages at scale 3 and scale 2 change to 7\% and 18\% respectively.

Using the chip generation algorithm, for a given minimum chip size (like $k=512$), we also compute the upper bound on the speedup which can be obtained. This is under the assumption that FocusPixels can be predicted without any error (\ie based on GTs). The bound for the speedup can change as we change the minimum chip size in the algorithm. Fig \ref{fig:upperbound} shows the effect of the minimum chip size parameter $k$ for FocusChip generation in algorithm \ref{alg:chip_generation}. The same value is used at each scale. For example, reducing the minimum chip size from 512 to 64 can lead to a theoretical speedup of $\sim10$ times over the baseline which performs inference on 3 scales. However, a significant reduction in minimum chip size can also affect detection performance - a reasonable amount of context is necessary for retaining high detection accuracy.

\begin{figure*}
    \centering
    \begin{subfigure}[t]{0.245\linewidth}
    \includegraphics[width=\linewidth]{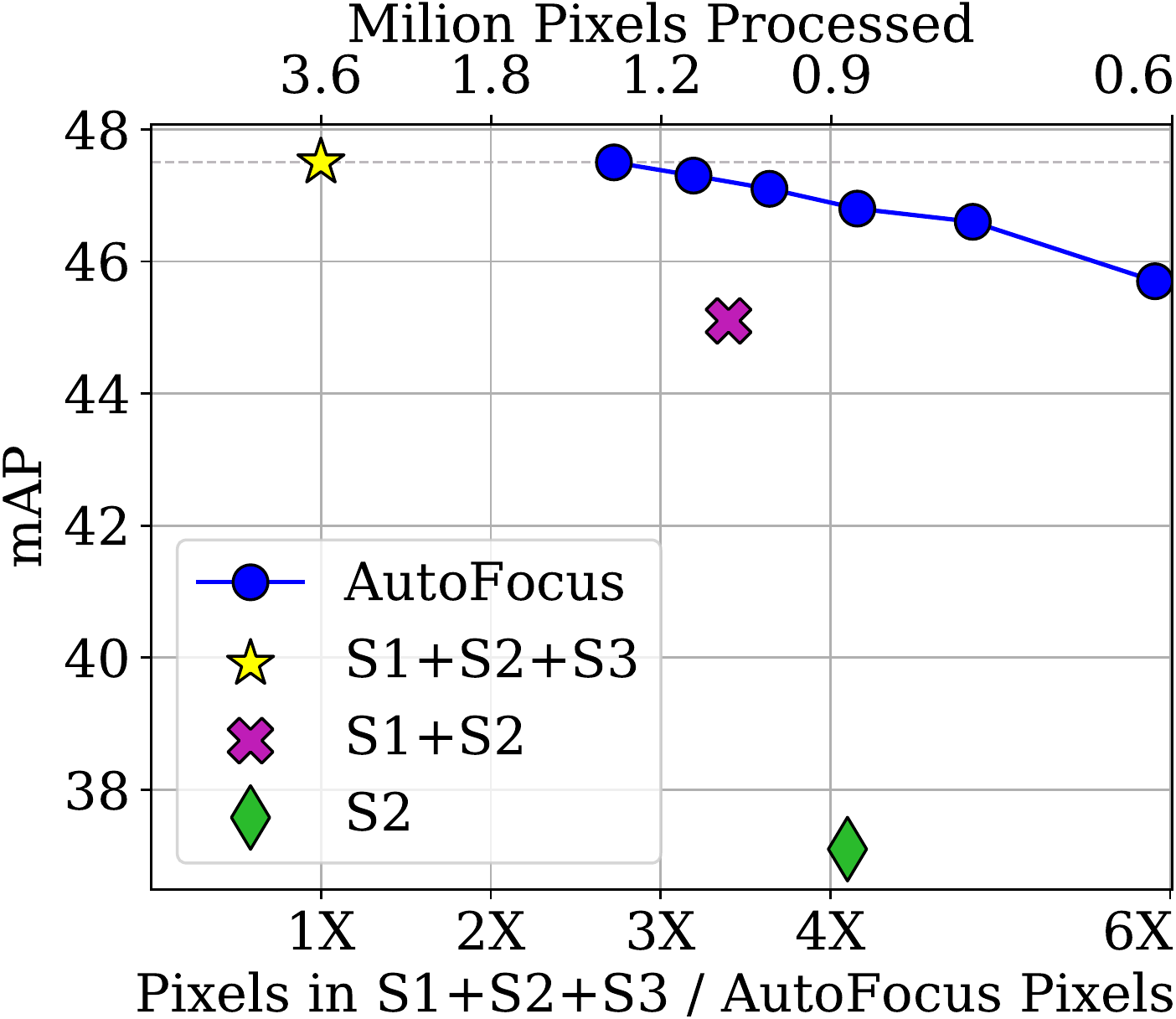}
    \caption{}
    \end{subfigure}
    \begin{subfigure}[t]{0.245\linewidth}
    \includegraphics[width=\linewidth]{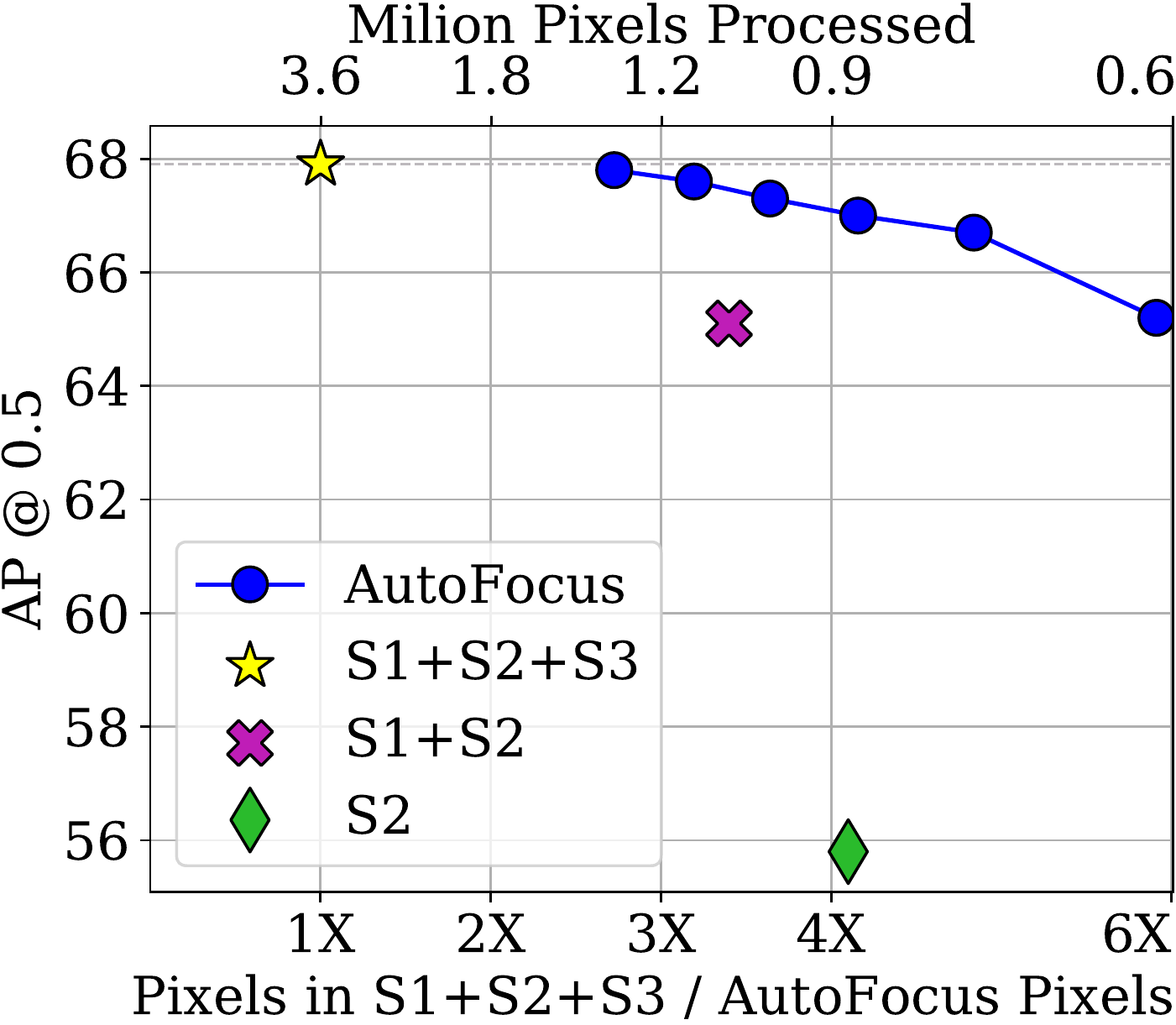}
    \caption{}
    \end{subfigure}
      \begin{subfigure}[t]{0.25\linewidth}
    \includegraphics[width=\linewidth]{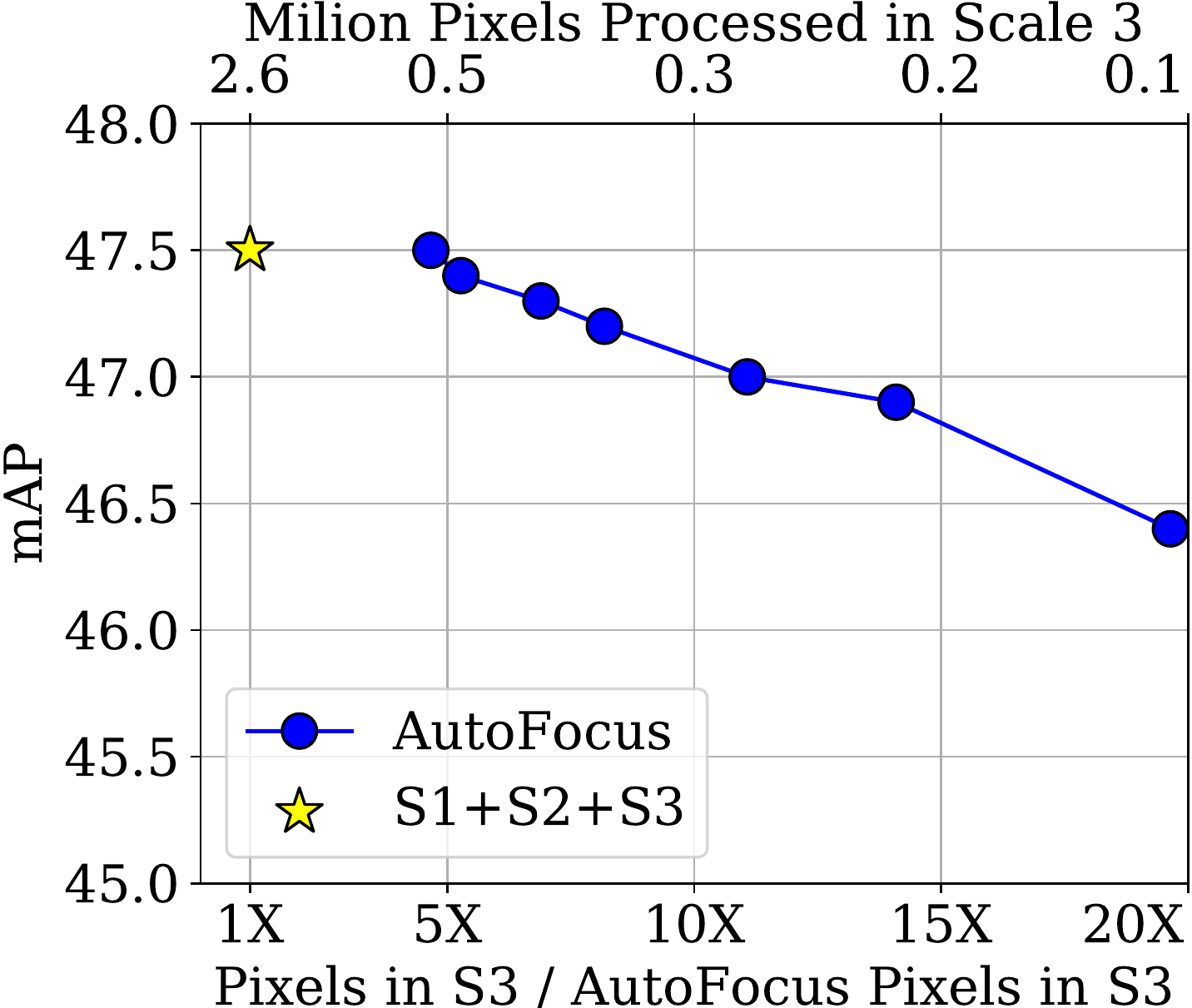}
    \caption{}
    \end{subfigure}
    \begin{subfigure}[t]{0.245\linewidth}
    \includegraphics[width=\linewidth]{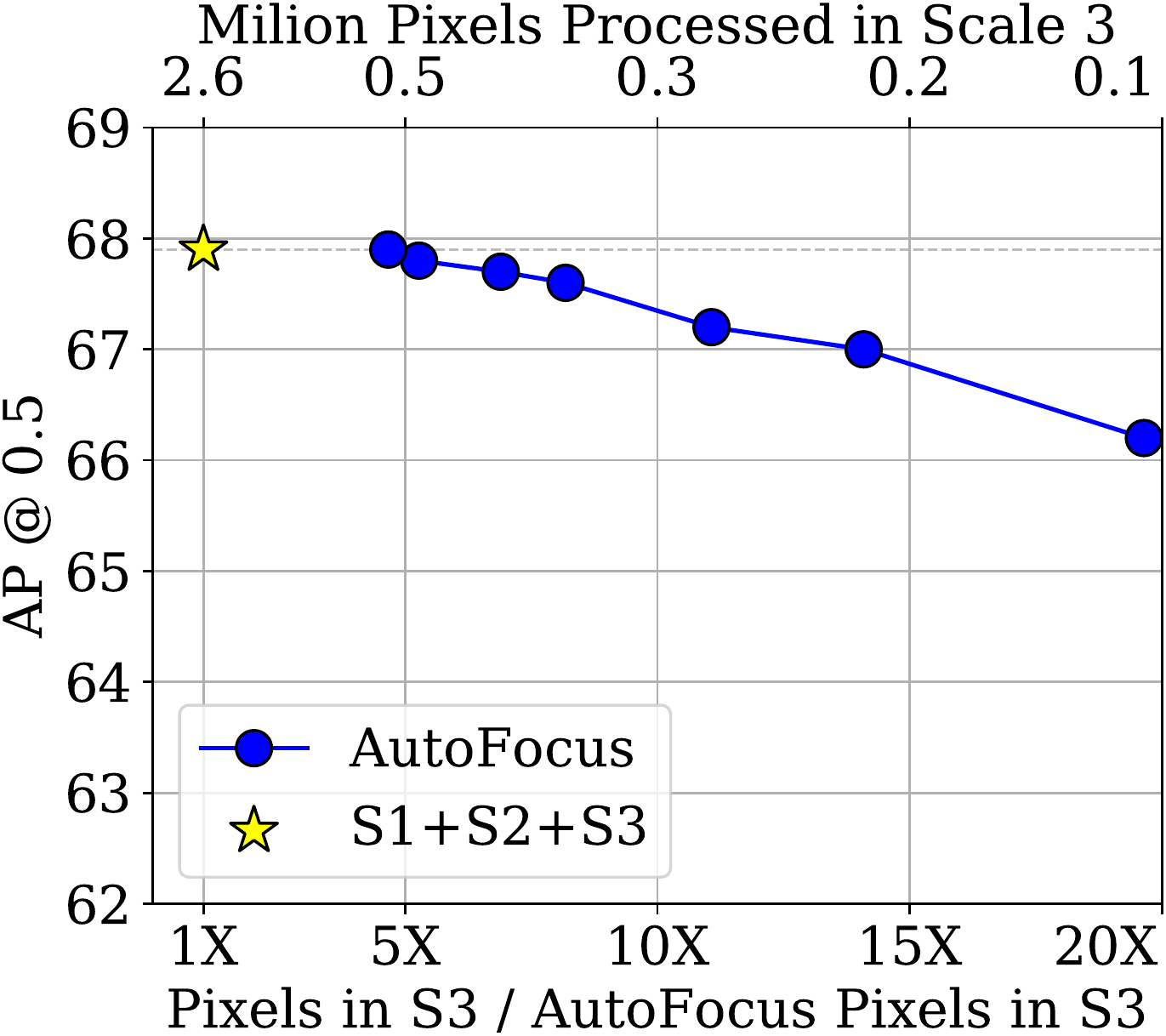}
    \caption{}
    \end{subfigure}
    \caption{Results are on the val-2017 set. (a,c) show the mAP averaged for IoU from 0.5 to 0.95 with an interval of 0.05 (COCO metric). (b,d) show mAP at 50\% overlap (PASCAL metric). We can reduce the number of pixels processed by a factor of 2.8 times without any loss of performance. A 5 times reduction in pixels is obtained with a drop of 1\% in mAP. }
    \label{fig:speed_acc}
\end{figure*}

\subsection{Quality of FocusPixel prediction}
\label{sec:focus_pixel_quality}
We evaluate how well our network predicts FocusPixels at different scales. To measure the performance, we use two criteria. First, we measure recall for predicting FocusPixels at two different resolutions. This is shown in Fig \ref{fig:my_label} \textcolor{red}{a}. This gives us an upper bound on how accurately we localize small objects using low resolution images. However, not all ground-truth objects which are annotated might be correctly detected. Note that our eventual goal is to accelerate the detector. Therefore, if we crop a region in the image which contains a ground-truth instance but the detector is not able to detect it, cropping that region would not be useful. The final effectiveness of FocusChips is coupled with the detector, hence we also evaluate the accuracy of FocusPixel prediction on regions which are confidently detected as shown in Fig \ref{fig:my_label} \textcolor{red}{b}. To this end, we only consider FocusPixels corresponding to those GT boxes which are covered (IoU $>$ 0.5) by a detection with a score greater than 0.5. At a threshold of 0.5, the detector still obtains an mAP of 47\% which is within 1\% of the final mAP and does not have a high false positive rate.

As expected, we obtain better recall at higher resolutions with both metrics. We can cover all confident detections at the higher resolution (scale 2) when the predicted FocusPixels cover just 5\% of total image area. At a lower resolution (scale 1), when the FocusPixels cover 25\% of the total image area, we cover all confident detections, see Fig \ref{fig:my_label} \textcolor{red}{b}. 

\subsection{Quality of FocusChips}
While FocusPixels are sufficient to generate enclosing regions which need to be processed, current software implementations require the input image to be a rectangle for efficient processing. To this end, we evaluate the performance of the enclosing chips generated using the FocusPixels. Similar to Section \ref{sec:focus_pixel_quality}, we use two metrics - one is recall of all GT boxes which are enclosed by FocusChips, the other one is recall for GT boxes enclosed by FocusChips which have a confident overlapping detection. To achieve perfect recall for confident detections at scale 2, FocusChips cover 5\% more area than FocusPixels. At scale 1, they cover 10\% more area. This is because objects are often not rectangular in shape. These results are shown in Fig \ref{fig:my_label} \textcolor{red}{d}.

\subsection{Speed Accuracy Trade-off}
We perform grid-search on different parameters, which are dilation, min-chip size and the threshold to generate FocusChips on a subset of 100 images in the validation set. For a given average number of pixels, we check which configuration of parameters obtains the best mAP on this subset. Since there are two scales at which we predict FocusPixels, we first find the parameters of AutoFocus when it is only applied to the highest resolution scale. Then we fix these parameters for the highest scale, and find parameters for applying AutoFocus at scale 2.

In Fig \ref{fig:speed_acc} we show that the multi-scale inference baseline which uses 3 scales obtains an mAP of 47.5\% (and 68\% at 50\% overlap) on the val-2017 set. Using only the lower two scales obtains an mAP of 45.4\%. The middle scale alone obtains an mAP of 37\%. This is partly because the detector is trained with the scale normalization scheme proposed in \cite{singh2017analysis}. As a result, the performance on a single scale alone is not very good, although multi-scale performance is high. The maximum savings in pixels which we can obtain while retaining performance is 2.8 times. We lose approximately 1\% mAP to obtain a 5 times reduction over our baseline in the val-2017 set.

We also perform an ablation experiment for the FocusPixels predicted using scale 2. Note that the performance of just using scales 1 and 2 is 45\%. We can retain the original performance of 47.5\% on the val-2017 set by processing just one fifth of scale 3. With a 0.5\% drop we can reduce the pixels processed by 11 times in the highest resolution image. This can be improved to 20 times with a 1\% drop in mAP, which is still 1.5\% better than the performance of the lower two scales.

\begin{table}[!ht]
\small
\begin{tabular}{c|c|c|c|c|c|c}
  \textbf{Method} & \textbf{Pixels} & \textbf{AP} & \textbf{AP}$^\textbf{50}$ & \textbf{S} & \textbf{M} & \textbf{L} \\
  \toprule
  Retina \cite{lin2018focal} & 950$^2$ & 37.8 & 57.5 & 20.2 & 41.1 & 49.2 \\
  LightH \cite{li2017light} & 940$^2$ & 41.5 & - & 25.2 & 45.3 & 53.1 \\
  Refine+ \cite{zhangsingle} & 3100$^2$ & 41.8 & 62.9 & 25.6 & 45.1 & 54.1 \\
  Corner+ \cite{law2018cornernet} & 1240$^2$ & 42.1 & 57.8 & 20.8 & 44.8 & 56.7 \\
  \midrule
 SNIPER \cite{sniper2018}  & 1910$^2$ & \textbf{47.9} & \textbf{68.3} & \textbf{31.5} & \textbf{50.5} & \textbf{60.3} \\ 
 \midrule
    & 1175$^2$ & \textbf{47.9} & \textbf{68.3} & \textbf{31.5} & \textbf{50.5} &  \textbf{60.3} \\  
  \textbf{AutoFocus}   & 930$^2$ & 47.2 & 67.5 & 30.9 & 49.0 & 60.0 \\
     & 860$^2$ & 46.9 & 67.0 & 30.1 & 48.9 & 60.0 \\
 \bottomrule
 \end{tabular}
\caption{Comparison with SNIPER on the COCO test-dev. This is our multi-scale baseline. 
 Results for others are taken from the papers/GitHub of the authors. Note that average pixels processed over the dataset are reported (instead of the shorter side). All methods use a ResNet-101 backbone. `+' denotes the multi-scale version provided by the authors.}
 \label{tab:res}
\end{table}

\begin{table}[!ht]
\small
\centering
\begin{tabular}{c|c|c|c}
  \textbf{Method} & \textbf{Pixels} & \textbf{AP}$^\textbf{50}$ & \textbf{AP}$^\textbf{70}$ \\
  \toprule
  Deformable ConvNet \cite{dai2017deformable} & 705$^2$ & 82.3 & 67.8\\
  Deformable ConvNet v2 \cite{zhu2018deformable} & 705$^2$ & 84.9 & 73.5\\
  \midrule

 SNIPER \cite{sniper2018}  & 1915$^2$ & 86.6 & 80.5 \\ 
 \midrule
\textbf{AutoFocus*}& 860$^2$ & 85.8 & 79.5 \\ 
 \midrule
\multirow{2}{*}{\textbf{AutoFocus}} & 700$^2$ & 85.3 & 78.1 \\
 & 1250$^2$ & 86.5 & 80.2 \\
    
 \bottomrule
 \end{tabular}
\caption{Comparison on PASCAL VOC 2007 test-set. All methods use ResNet-101 and trained on VOC2012 trainval+VOC2007 trainval. The average pixels processed over the  dataset  are  also reported. To show the robustness of AutoFocus to hyper-parameter choices, in `*' we use the same parameters as COCO and run the algorithm on PASCAL.}
\vspace{-2mm}
 \label{tab:res_voc}
\end{table}

\begin{figure*}[!ht]
    \centering
    \includegraphics[width=\linewidth]{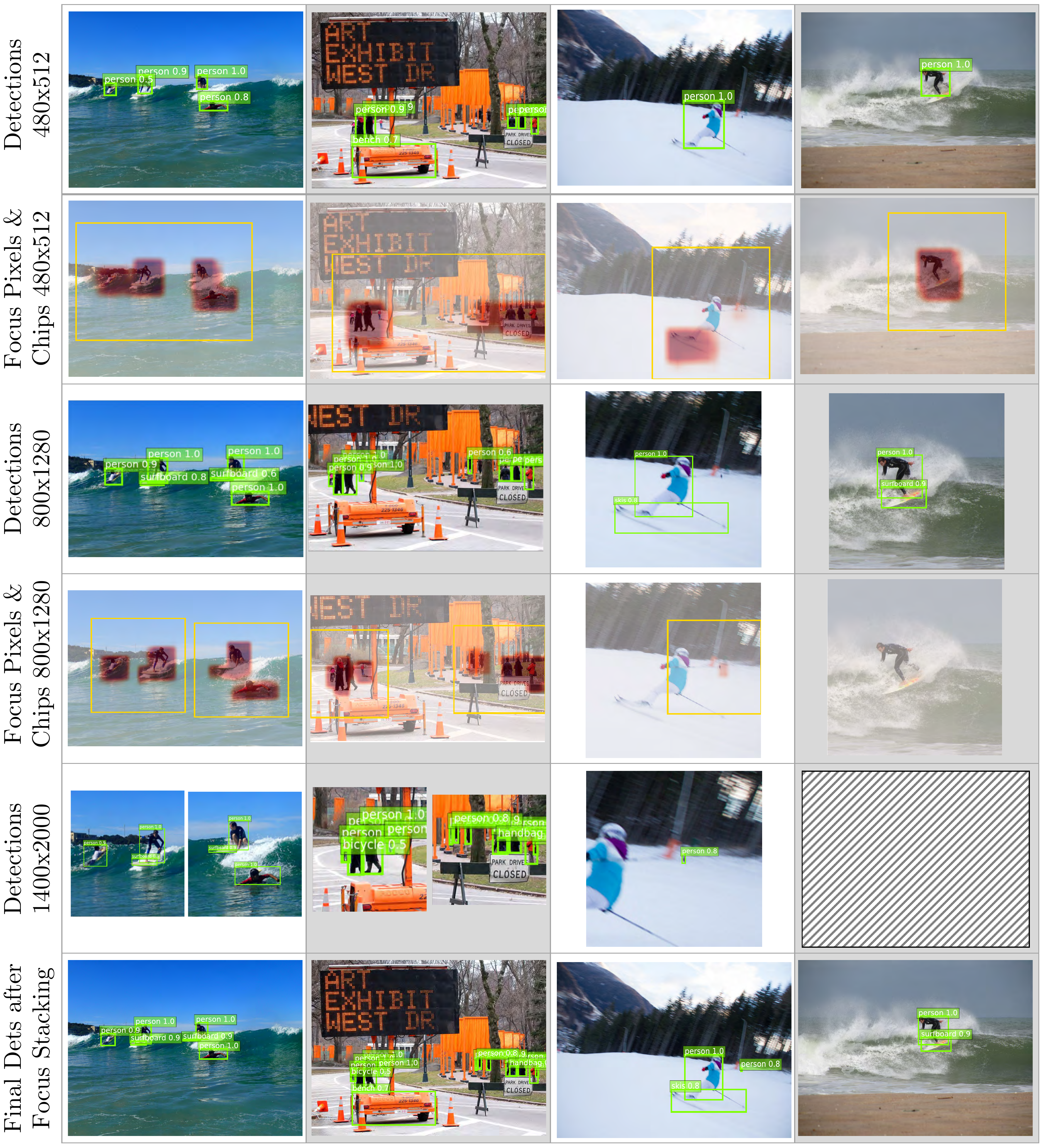}
    \caption{Each column shows the inference pipeline in AutoFocus. The confidence for FocusPixels and FocusChips are shown in red, and yellow respectively in the second and fourth rows. Detections are shown in green. As can be seen, complex images containing many small objects like the two leftmost columns can generate multiple FocusChips in high resolutions like $1400\times2000$. Images which do not contain small objects are not processed at all in high resolution, like the one in the rightmost column.}
    \label{fig:quals}
\end{figure*}

Results on the COCO test-dev set are provided in Table \ref{tab:res}. \textbf{While matching SNIPER's performance of 47.9\% (68.3\% at 0.5 IoU), AutoFocus processes 6.4 images per second on the test-dev set with a Titan X Pascal GPU.} SNIPER processes 2.5 images per second. RetinaNet with a ResNet-101 backbone and a FPN architecture processes 6.3 images per second on a P100 GPU (which is like Titan X), but obtains 37.8\% mAP \footnote{\url{https://github.com/facebookresearch/Detectron/blob/master/MODEL_ZOO.md}}. We also report the number of pixels processed with a few efficient recent detectors. Detectors which perform better than SNIPER like MegDet \cite{peng2017megdet} or PANet \cite{liu2018path} are slower because they use complex architectures like ResNext-152 \cite{xie2017aggregated} etc. To the best of our knowledge, AutoFocus is the fastest detector which obtains an mAP of 47.9\% (or 68.3\% at 0.5 IoU) on the COCO dataset. We show the inference process for AutoFocus on a few images in the COCO val-2017 set in Fig \ref{fig:quals}. We also report results on the PASCAL VOC dataset in Table \ref{tab:res_voc}. To show the robustness of AutoFocus to its hyper-parameters, we use exactly the same hyper-parameters tuned for COCO (shown as AutoFocus*). While processing the same area as DeformableV2 \cite{zhu2018deformable}, AutoFocus achieves 4.6\% better AP at 0.7 IoU. It also matches the performance of SNIPER while being considerably more efficient. Its mAP (on the COCO metric) can be further improved by using refinement techniques like cascade-RCNN \cite{cai2017cascade}. 

\section{Future Work}

While results for only a Faster R-CNN based detector were presented, it is possible to combine AutoFocus with detectors like YOLOv2 \cite{redmon2017yolo9000}, RetinaNet \cite{lin2018focal} and for other instance-level recognition tasks like pose estimation, instance segmentation \etc. It would also be of interest to develop efficient multi-scale algorithms for tasks like stuff segmentation. Combining tracking and multi-scale inference can lead to further acceleration in videos.

\vspace{2mm}
\textbf{Acknowledgement}
The research is based upon work supported by the Office of the Director of National Intelligence (ODNI), Intelligence Advanced Research Projects Activity (IARPA), via DOI/IBC Contract Numbers D17PC00287 and D17PC00345. The U.S. Government is authorized to reproduce and distribute reprints for Governmental purposes not withstanding any copyright annotation thereon. Disclaimer: The views and conclusions contained herein are those of the authors and should not be interpreted as necessarily representing the official policies or endorsements, either expressed or implied of IARPA, DOI/IBC or the U.S. Government. The authors would also like to thank an Amazon Machine Learning gift for the AWS credits used for this research.

{\small
\bibliographystyle{ieee}
\bibliography{main}
}

\end{document}